# Color Texture Classification Approach Based on Combination of Primitive Pattern Units and Statistical Features

Shervan Fekri Ershad

Department of computer science, engineering and IT, Shiraz International University, Shiraz, Iran
*shfekri@shirazu.ac.ir*

## Abstract

*Texture classification became one of the problems which has been paid much attention on by image processing scientists since late 80s. Consequently, since now many different methods have been proposed to solve this problem. In most of these methods the researchers attempted to describe and discriminate textures based on linear and non-linear patterns. The linear and non-linear patterns on any window are based on formation of Grain Components in a particular order. Grain component is a primitive unit of morphology that most meaningful information often appears in the form of occurrence of that. The approach which is proposed in this paper could analyze the texture based on its grain components and then by making grain components histogram and extracting statistical features from that would classify the textures. Finally, to increase the accuracy of classification, proposed approach is expanded to color images to utilize the ability of approach in analyzing each RGB channels, individually. Although, this approach is a general one and it could be used in different applications, the method has been tested on the stone texture and the results can prove the quality of approach.*

## 1. Introduction

Texture classification has been paid much attention on by Image processing and computer vision scientists since 1980s. As texture classification has a close relationship with the sciences such as machine learning and artificial Intelligence. It functions in areas like pattern recognition, object tracking, defect detection, face tracking, Image Segmentation, Image retrieval and etc. Researchers try to offer methods to do texture classification with high accuracy by analyzing texture. So far, many methods have been offered to solve this problem, such as Texture classification based on random threshold vector [1], Texture classification by statistical learning from morphological image processing [2], Texture Classification and Defect Detection by Statistical Features [7], Extracting of shape components for classification of textures based on texture elements [6], A Combined Color, Texture and Edge Features Based Approach for Identification and Classification [5]. The techniques used to texture analysis or classification are discussed in four categories, statistical approaches, Structural approaches filter based methods, and model based approaches. Table 1 shows a summary list of some of the key texture analysis methods that have been applied to Texture classification or segmentation. Clearly, statistical and filter based approaches have been very popular.

Statistical texture analysis methods measure the spatial distribution of pixel values. They are well rooted in the computer vision world and have been extensively applied to various tasks. A large number of statistical texture features have been proposed, ranging from first order statistics to higher order statistics. Amongst many, histogram statistics, co-occurrence matrices, autocorrelation, and local binary patterns have been applied to texture Analysis or Classification. In structural approaches, texture is characterized by *texture primitives* or texture elements, and the spatial arrangement of these primitives. Thus, the primary goals of structural approaches are firstly to extract texture primitives, and secondly to model or generalize the spatial placement rules. The texture primitive can be as simple as individual pixels, a region with uniform gray levels, or line segments. The placement rules can be obtained through modeling geometric relationships between primitives or learning statistical properties from texture primitives. The filter based techniques largely share a common characteristic, which is applying filter banks on the image and compute the energy of the filter responses. The methods can be divided into spatial domain, frequency domain, and joint spatial/spatial-frequency domain techniques. Model based methods include, among many others, fractal models, autoregressive models, random field models, the epitome model, and the texem model.

In most approaches, which have been offered so far, researchers have tried to analyze and describe texture based on linear and non-linear patterns. In [3], Suresh & Kumar proposed a new method to analyze texture. In [3], a new concept called "Grain component" was put forward.

The algorithm, which has been proposed in this article for texture classification, in the first stage to analyze texture, has used the grain component concept. In the next stage, by measuring grain components, Grain components histogram has been defined and made. And, in the continuation, Statistical features have been extracted of histogram and an identification vector has been defined for every image to do texture classification. To increase the texture classification quality and the accuracy of the method, the color image of the texture has been divided to 3 sub-images in the color channels of R, G, and B and the proposed approach has been expanded to color images. The algorithm, which has been proposed in this article has not only a high speed in measurements, but also because of the texture images in every color channel individually, has done texture classification with a high accuracy. In the result section, to prove this claim, by gathering some images of textures in 4 models of stones, and applying proposed approach on them, stone texture classification has been done. The high accuracy in texture classification in the results shows the quality of offered approach.

| Category | Method |
|---|---|
| **Statistical** | 1. Histogram properties<br>2. Co-occurrence matrix<br>3. Local binary pattern<br>4. Other gray level statistics<br>5. Autocorrelation<br>6. Registration-based |
| **Structural** | 1. Primitive measurement<br>2. Edge Features<br>3. Skeleton representation<br>4. Morphological operations |
| **Filter Based** | 1. Spatial domain filtering<br>2. Frequency domain analysis<br>3. Joint spatial/spatial-frequency |
| **Model Based** | 1. Fractal models<br>2. Random field model<br>3. Texem model |

Table 1: Inexhaustive list of textural analysis methods

## 1.1 Paper Organization

The reminder of this paper is organized as follows: Section two is related to the description of Primitive pattern units (PPU) and the way of this estimation. Section three is related to the description of building histogram of PPUs and feature extraction. Section four has an algorithm for computing the feature vector for color images. and finally, the results and conclusion included.

## 2. Primitive Pattern Units

In [3] the authors said that, the texture is characterized not only by gray value at a given pixel, but also by the gray value pattern in the surrounding pixels. Correspond to this description, so the texture has both local and global meaning, in the sense that it is characterized by the invariance of certain local attributes that are distributed over a region of an image. Based on this, the present paper proposes a novel design approach for texture classification based on number of grain components on a 3×3 mask. A grain component is a Primitive Pattern Unit (PPU) of morphology, because of which most significant and meaningful information of a texture often appears in the form of occurrence of pixels on a neighborhood. That's why the present study used sum of occurrence of PPU's for feature extraction and there by classification.

The frequency of occurrence of PPU's are calculated in the following way. A PPU is counted if and only if the central pixel of the window is a grain. If the central pixel is not a grain then the window is treated as a zero grain component window which is shown in Fig.1. In the following figures '0' indicates no grain, '1' indicates a grain and "d" indicates don't care i.e. it can be either 0 or 1. There can be 8 combinations of PPU with one grain which are shown in the Fig. 2.

| d | d | d |
|---|---|---|
| d | 0 | d |
| d | d | d |

Fig.1. The possible zero grain components.

| 1 | 0 | 0 |     | 0 | 1 | 0 |
|---|---|---|-----|---|---|---|
| 0 | 1 | 0 |     | 0 | 1 | 0 |
| 0 | 0 | 0 |     | 0 | 0 | 0 |
|   (a)     |     |   (b)     |

| 0 | 0 | 1 |     | 0 | 0 | 0 |
|---|---|---|-----|---|---|---|
| 0 | 1 | 0 |     | 0 | 1 | 1 |
| 0 | 0 | 0 |     | 0 | 0 | 0 |
|   (c)     |     |   (d)     |

| 0 | 0 | 0 |     | 0 | 0 | 0 |
|---|---|---|-----|---|---|---|
| 0 | 1 | 0 |     | 0 | 1 | 0 |
| 0 | 0 | 1 |     | 0 | 1 | 0 |
|   (e)     |     |   (f)     |

| 0 | 0 | 0 |     | 0 | 0 | 0 |
|---|---|---|-----|---|---|---|
| 0 | 1 | 0 |     | 1 | 1 | 0 |
| 1 | 0 | 0 |     | 0 | 0 | 0 |
|   (g)     |     |   (h)     |

Fig.2. Representation of PPU with one grain

There will be 7 different formations of PPU's with two grain components by fixing one of the grains at pixel location (0,0) on a 3×3 mask as shown in Fig. 3. In the similar way there will be six formations of PPU with two grains by positioning one of the grains at the pixel location (0,1) as shown in Fig.4.

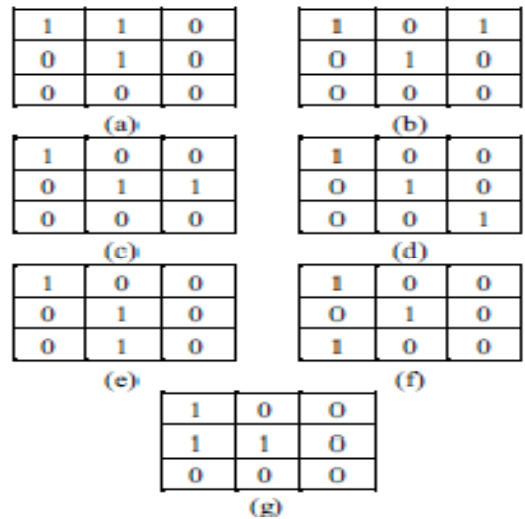

Fig.3. Representation of PPU with two grain components by fixing one of the grain component at (0, 0).

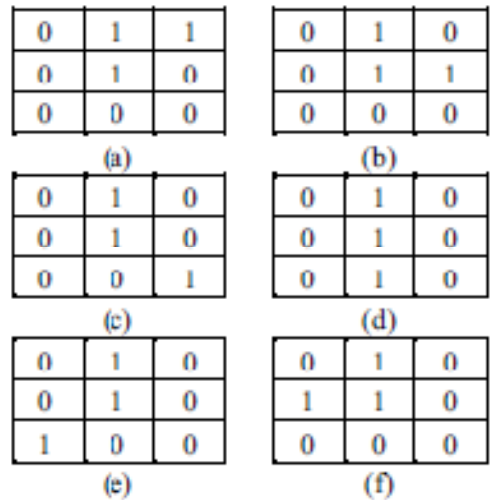

Fig.4 Representation of PPU with two grain components by fixing one of the grain component at (0, 1).

There will be 7! ways of forming PPU's with two grain components for a 3×3 window. In the same way there will be 6!, 5!, 4!, 3!, 2! and 1! ways of forming PPU with 3, 4, 5, 6, 7 and 8 grain components respectively.. The frequency of occurrences of PPU's from one to eight grains are computed by the proposed algorithm on a 3×3 non overlapped mask for all stone textures. The Tables 2 represents the frequency of occurrences of PPU's for some of the stone textures respectively that will use in result part (the stone images converted to gray-level to compute the frequencies). It's important to say that, to compute the frequency of occurrence, there is thresholding stage. All of the thresholding methods like Valley-emphasis can be useable in this method but for increase the accuracy, Otsu method has been used in this paper.

| Image | Grain Components | | | | |
|---|---|---|---|---|---|
| | 0 | 1 | 2 | 3 | 4 |
| Granite1 | 3245 | 76 | 121 | 245 | 401 |
| Granite2 | 3320 | 54 | 117 | 301 | 382 |
| Granite3 | 3214 | 64 | 119 | 343 | 403 |
| Teravertine1 | 4187 | 48 | 97 | 356 | 443 |
| Teravertine2 | 4021 | 44 | 107 | 445 | 460 |
| Teravertine3 | 4004 | 49 | 112 | 399 | 471 |
| Marble 1 | 3352 | 87 | 162 | 529 | 572 |
| Marble 2 | 3512 | 84 | 156 | 563 | 525 |
| Marble 3 | 3408 | 79 | 160 | 575 | 544 |
| Hatchet 1 | 4187 | 110 | 76 | 710 | 682 |
| Hatchet 2 | 4443 | 89 | 45 | 691 | 545 |
| Hatchet 3 | 4447 | 93 | 62 | 690 | 633 |

| Image | Grain Components | | | |
|---|---|---|---|---|
| | 5 | 6 | 7 | 8 |
| Granite 1 | 622 | 349 | 821 | 1345 |
| Granite 2 | 550 | 388 | 814 | 1299 |
| Granite 3 | 530 | 392 | 802 | 1348 |
| Teravertine1 | 283 | 441 | 623 | 747 |
| Teravertine2 | 301 | 712 | 575 | 560 |
| Teravertine3 | 304 | 722 | 601 | 573 |
| Marble 1 | 434 | 527 | 532 | 1030 |
| Marble 2 | 410 | 478 | 521 | 976 |
| Marble 3 | 414 | 483 | 562 | 1000 |
| Hatchet 1 | 320 | 335 | 295 | 510 |
| Hatchet 2 | 356 | 401 | 210 | 445 |
| Hatchet 3 | 336 | 377 | 214 | 517 |

Table 2. The frequency of occurrence of grain components for some of the result textures in Gray-levels

## 3. Building Histogram & Feature Extraction

As it was pointed out in the previous section, the occurrence frequent of each primitive pattern units has been computed for each image. Now, the histogram of Grain components can be built up in each image. In this way, every grain component in the histogram is a bin and frequency of occurrence of them, is the height of that bin. This is observable in figure 5. There, the occurrence frequency of each grain component has been computed for three images which are Granite1, Teravertine3 and Hatchet2, and then, theirs histograms has been drawn.

Now, by building grain components histogram for the texture image, different statistical features, such as Energy, Entropy can be computed for histogram. The measurable Statistical features have been shown in the equations 1to4. Finally, after measuring the amount of statistical features, an identity feature vector called "F" can be defined for every image like F= <Energy, Entropy, Mean,

Variance > having four dimensions and the value of every dimension is equal to measured statistical features.

$$Mean = \bar{g} = \sum_{g=0}^{L} gP(g) \tag{1}$$

$$Variance = \sigma_g = \sqrt{\sum_{g=0}^{L} (g - \bar{g})^2 P(g)} \tag{2}$$

$$Energy = \sum_{g=0}^{L} [P(g)]^2 \tag{3}$$

$$Entropy = -\sum_{g=0}^{L} P(g) \log_2 [P(g)] \tag{4}$$

Where P (g) is probability of grin component "g" in the image, So P(g) is computable by (5), it means the normalization of Grain component Histogram. Also according to the section 2, range of "g" is 0 to 8 for 3*3 masks. Consequently L is 8. For another size of masks like 5*5 or 7*7 or etc, there is a same way. The number of grain components just may increase but the dimension of feature vector is fixed. for example for 5*5 mask there is 25 form of grain components. Non-sensitivity to size of masks is one of advantages of the proposed approach.

P (g) = N(g)/M        (5)

Where N(g) is the frequency of occurrence of grain component "g" and M is the sum of the all grain components occurrence frequent. Some of the other statistical features like contrast, homogeneity and correlation can use, but the results shows this features cannot provide a good discrimination. For example the feature vector F1 is computed for Marble2, which is shown in equation (6).

Energy = 0.271    Entropy = 2.327    Mean = 2.68    Variance = 3.08
F1 = <0.271, 2.327, 2.68, 3.08>        (6)

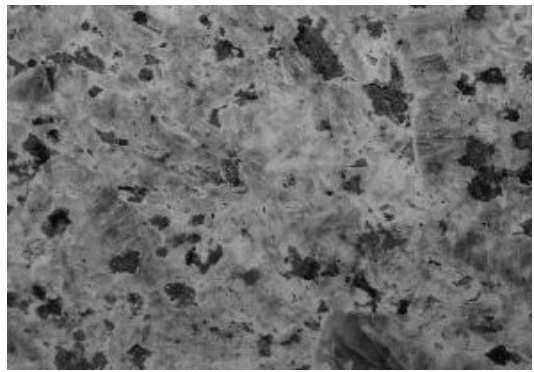

(a)

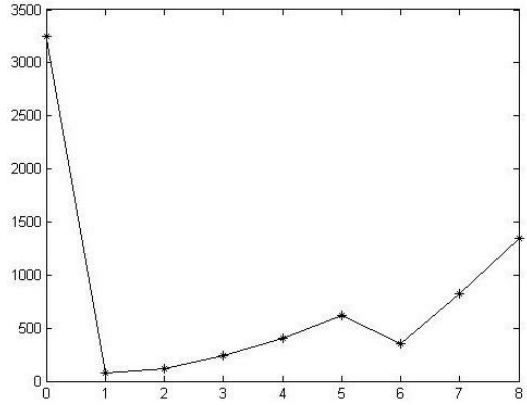

(b)

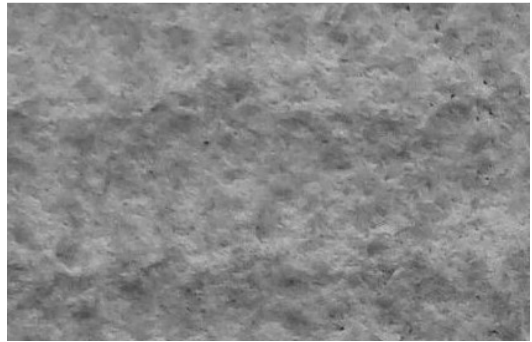

(c)

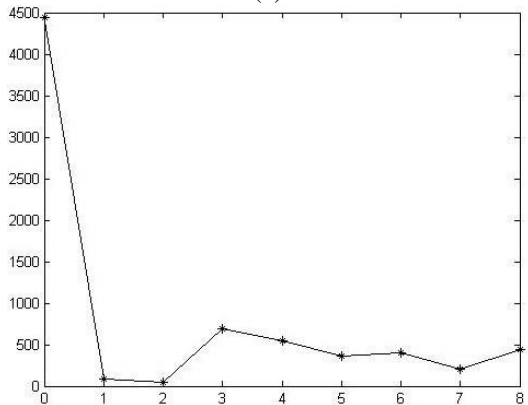

(d)

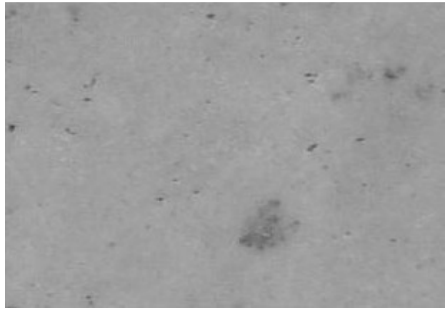

(e)

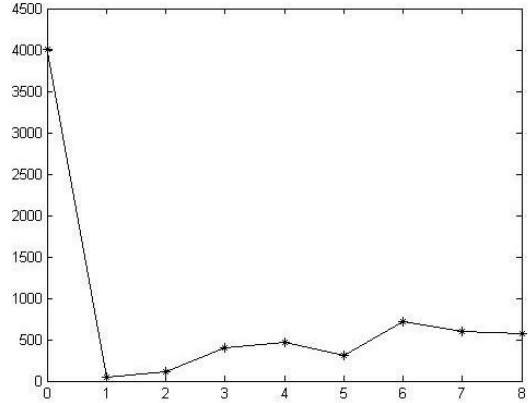

(f)

Fig 5. (a) Original Image of Granite 1(b) Grain Components Histogram of Granite 1 (c) Original Image of Hatchet 2(d) Grain Components Histogram of Hatchet 2 (e) Original Image of Teravertine3 (f) Grain Components Histogram of Teravertine3

## 4. Color Texture analysis & Classification

The approach which was explained in sections 2 and 3 can be used for Gray-level Texture classification. In this section, an approach was explained that can be expanded to the color texture images to increase the texture classification accuracy. In [4], X.Maldague and A. Akhloufi offered an approach to analyze the color images based on analyzing each color channels individually. In this article too, the same approach has been used to expand it to the color images. The space of the color images is usually RGB or they are capable of turning into RGB's space. Each color image in the space of RGB is consisting of 3 color channels of Red, Green and Blue. and, every pixel in each channel has a special intensity which makes a color pixel when being mixed. So, for every color channel, the frequency of occurrence of primitive pattern units can be computed individually by using the algorithm proposed in section2. The Tables 3 represents the frequency of occurrences of PPU's in each RGB channels individually for some of the stone color textures respectively that will use in result part.

In continuation, for every color channels, a grain components histogram is buildable. After making histograms and extracting statistical features of them, based on the algorithm proposed in section 3, three feature vectors are provided. Finally, by linking 3 feature vectors, a feature vector of 12dimension is provided for each texture image. Equation (7) shows this concept.

$$F_{RED} = \langle Energy_{HR}, Entropy_{HR}, Mean_{HR}, Variance_{HR} \rangle$$
$$F_{Green} = \langle Energy_{HG}, Entropy_{HG}, Mean_{HG}, Variance_{HG} \rangle$$
$$F_{Blue} = \langle Energy_{HB}, Entropy_{HB}, Mean_{HB}, Variance_{HB} \rangle$$
$$F_{total} = \langle F_{red}, F_{green}, F_{blue} \rangle =$$
$$< Energy_{HR}, Entropy_{HR}, Mean_{HR}, Variance_{HR}, Energy_{HG}, Entropy_{HG}, Mean_{HG}, Variance_{HG},$$
$$Energy_{HB}, Entropy_{HB}, Mean_{HB}, Variance_{HB} > \qquad (7)$$

Where HR is grain components histogram of color texture in red channel, also HG means histogram of grain components in green channel. HB is the same histogram in blue channel. The feature vector, computed by this approach for every color texture, is a very good characteristic for the label of that texture. Of different advantages in this feature extraction, facility of computing and adapting to different classifiers can be mentioned.

Now, in order to do texture classification, it's enough to use some training images to make dataset of the textures for each label. Now, classifiers are used in the test data to do the classification. According to the introduction, the proposed approach can group in Statistical category and Model based category. Analysis the image by primitive pattern units and compute the amount of grain components means analysis the texture based on its texems so it can group in model based approaches. Also computing the statistical feature for grain components histogram groups the proposed approach in statistical category. Grain components histogram of granite 1in red, green and blue channel has shown in figure6 individually.

### 4.1. Preprocessing

To increase the quality of texture classification a preprocessing stage is need. Preprocessing should normal the histogram of intensities and decreases the illumination. So histogram equalization method is used. Histogram equalization transforms an image with an arbitrary histogram to one with a flat histogram.

| Image | Grain Components for Red Channel | | | | | | | | |
|---|---|---|---|---|---|---|---|---|---|
| | 0 | 1 | 2 | 3 | 4 | 5 | 6 | 7 | 8 |
| **Granite1** | 3225 | 67 | 132 | 298 | 445 | 710 | 320 | 881 | 1147 |
| **Teravertine1** | 4107 | 52 | 98 | 342 | 491 | 628 | 390 | 720 | 397 |
| **Marble 1** | 3751 | 91 | 172 | 497 | 563 | 439 | 483 | 631 | 598 |
| **Hatchet 1** | 4554 | 109 | 118 | 518 | 631 | 323 | 502 | 263 | 207 |

| Image | Grain Components for Green Channel | | | | | | | | |
|---|---|---|---|---|---|---|---|---|---|
| | 0 | 1 | 2 | 3 | 4 | 5 | 6 | 7 | 8 |
| **Granite1** | 3177 | 62 | 102 | 301 | 453 | 681 | 324 | 795 | 1330 |
| **Teravertine1** | 3882 | 73 | 87 | 372 | 522 | 632 | 347 | 801 | 509 |
| **Marble 1** | 3641 | 81 | 203 | 483 | 572 | 399 | 461 | 671 | 714 |
| **Hatchet 1** | 4232 | 119 | 128 | 487 | 701 | 326 | 563 | 257 | 412 |

| Image | Grain Components for Blue Channel | | | | | | | | |
|---|---|---|---|---|---|---|---|---|---|
| | 0 | 1 | 2 | 3 | 4 | 5 | 6 | 7 | 8 |
| **Granite1** | 3065 | 63 | 332 | 442 | 473 | 655 | 192 | 1022 | 981 |
| **Teravertine1** | 3640 | 55 | 83 | 331 | 500 | 523 | 470 | 1188 | 435 |
| **Marble 1** | 4001 | 82 | 133 | 467 | 551 | 420 | 460 | 541 | 570 |
| **Hatchet 1** | 4054 | 111 | 123 | 509 | 726 | 328 | 507 | 244 | 623 |

Table 3. The frequency of occurrence of grain components for some of the result textures in each RGB channels individually

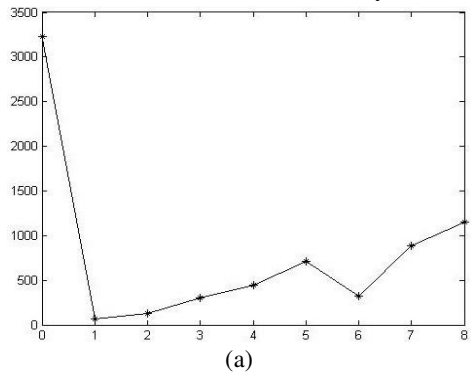

(a)

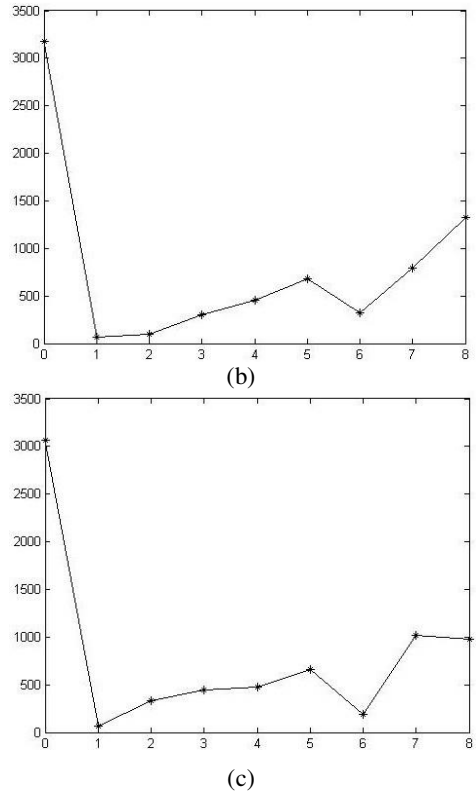

(b)

(c)

Fig 6. (a) Grain Components Histogram of Granite 1 in red channel (b) Grain Components Histogram of Granite 1 in green channel (c) Grain Components Histogram of Granite 1in blue channel

## 5. Results

The approach offered in this article has got a practical and general aspect. It can be used to do classification on every kind of texture perfectly. To survey the quality of the proposed approach, stone texture classification was done. First, 80 high resolution color images were taken from different stone models named "granite" & "travertine" & "hatchet" and "marble" by a digital camera.

It means 20 images for every stone model. Then divided 52 images for the train stage, and, 28 images for the test stage. In continuation, the proposed approach was applied on the train images and a dataset was made with 52 samples, each of samples which having 12 dimensions, and each samples having a label, which is the model of that stone. Finally, getting help from classifiers, such as KNN, Decision Table, J48tree and Naïve beyes, the 32 test images were classified. Tables 4 & 5 show the accuracy rate of classifications. Also, using two other methods, like local binary pattern (LBP) and gray level co-occurrence matrix (GLCM), the classification was done. LBP and GLCM are describing some new features to texture analysis. After that, texture classification done by using classifiers. As it is seen, the accuracy rate in the proposed approach has been higher than in the previous ones. And, the dataset has got a high adaptation to the kinds of classifiers. That is an advantage for the texture analysis approach which is proposed in this article. Also introducing useful series of features and low computation complexity are some of the other advantages.

| Classifiers  Approach | 1NN : (Accuracy ± std-dev) | 3NN : (Accuracy ± std-dev) | 5NN : (Accuracy ± std-dev) |
|---|---|---|---|
| **Proposed Approach** | **93.31± 0.5** | **95.23 ±0.5** | **94.31±0.4** |
| LBP | 91.97±0.2 | 92.54±0.4 | 92.17±0.2 |
| GLCM | 89.24±0.3 | 88.65±0.7 | 86.54±0.5 |

Table 4. Accuracy rate of classifiers on methods

| Classifiers  Approach | Decision Table : (Accuracy ± std-dev) | J48Tree : (Accuracy ± std-dev) | Naïve Bayes : (Accuracy ± std-dev) |
|---|---|---|---|
| **Proposed Approach** | **89.78± 0.5** | **92.65 ±0.2** | **94.20±0.3** |
| LBP | 87.43±0.6 | 91.82±0.3 | 89.60±0.3 |
| GLCM | 82.27±0.6 | 84.72±0.7 | 85.23±0.5 |

Table 5. Accuracy rate of classifiers on methods

## 6. Conclusion

The purpose of this article is offering an approach for texture classification. In this respect, in section 2, by using the concept of primitive pattern units, the texture has been analyzed. After that, by building histogram of grain components and extracting statistical features from that, the feature extraction has been done. So, a computable feature vector which has most meaningful information of texture is provided. Then the dataset was made in the train stage by using feature vectors which has been extracted from texture images. So, by using classifiers, texture classification was done in test stage. In the continuation, the approach was expanded to the color texture images. In the result section, the proposed approach was applied on the images of the 4 kinds of stones and the image texture was analyzed and the classification was done.

Also, for comparison, the images were analyzed and classified by 2 kind of previous methods name local binary pattern and gray level co-occurrence matrix. The results at the end show that the approach which has been proposed in this article has got a high ability and accuracy in texture classification. The main advantages of the proposed approach in this article can be mentioned in two points. 1-Corresponding with near all of the classifiers, 2-Introducing a series of new features which are able to be computed for different applications.

## 7. Future Work

One interesting future research direction is to use the proposed approach to texture segmentation or object tracking by analyzing the object or image via feature vector $F_{total}$. Also future research direction is to use another color spaces such as HSV or YCbCr and compare the results with RGB space.

## Acknowledgement

The Author likes to say thanks to Mr. Mohammad Abdollahi & Mrs. Sepideh Fekri Ershad (both of them, Master Graduated of Civil Engineering of AmirKabir University of Iran) for their Guides to collate the requirement database. Also special thanks to Mr. Arzhang Seyrafi (Master student of Computer Science in Turin University of Italy) for his supports.

## Author

**Shervan Fekri Ershad** has M.S. degree in Artificial Intelligence from International Shiraz University of Iran. His license degree is in Hardware computer engineering from Islamic Azad university of Najaf Abad. His research areas include image processing, visual pattern recognition, Inspection systems, texture Analysis& classification and object tracking.
.

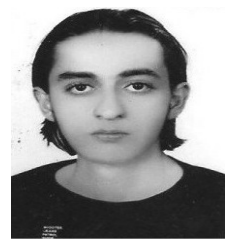